\documentclass{article}
\usepackage{ijcai26}

\usepackage{times}
\usepackage{soul}
\usepackage{url}
\usepackage[hidelinks]{hyperref}
\usepackage[utf8]{inputenc}
\usepackage[small]{caption}
\usepackage{graphicx}
\usepackage{amsmath}
\usepackage{amsthm}
\usepackage{booktabs}
\usepackage{algorithm}
\usepackage{algorithmic}
\usepackage[switch]{lineno}
\usepackage{array}
\usepackage{xcolor}
\usepackage{booktabs}
\usepackage{multirow}
\usepackage{graphicx}
\usepackage{amsfonts}

\title{RetiSEM: Generalising Causal Models for Fragmented Biomedical Data}

\author{
Inam Ullah$^{1*}$
\and
Imran Razzak$^2$
\and
Shoaib Jameel$^1$
\\
\affiliations
$^1$University of Southampton, Southampton, SO18 1PB, Hampshir, United Kingdom\\
$^2$Mohamed bin Zayed University of Artificial Intelligence, United Arab Emirates
\\
\emails
i1n23@soton.ac.uk,
Imran.razzak@mbzuai.ac.ae,
M.S.Jameel@southampton.ac.uk
}

\begin{document}

\maketitle

\begin{abstract}
Learning causal models from fragmented biomedical data is challenging because clinical, molecular, and imaging variables are often incomplete or not jointly observed. We propose RetiSEM, a domain-constrained structural equation modelling (SEM) framework for causal graph recovery and mediation analysis under limited multimodal resources. This proposed work organises variables into biologically informed blocks, applies forbidden-edge constraints, and decomposes pathway-level effects into TE, NDE, and NIE components. We evaluate RetiSEM across ten synthetic benchmark scenarios that vary in dimensionality, nonlinearity, causal depth, and pathway structure, together with a fragmented real-world setting that combines NHANES clinical variables with externally derived retinal representations. This approach achieves lower structural error and higher causal accuracy than unconstrained baselines across the synthetic benchmarks. In the real-data analysis, retinal variables behave mainly as downstream biomarker-like indicators, with smaller but detectable indirect effects. These findings support our strategy as an interpretable framework for testing structured causal hypotheses in limited-resource biomedical AI. The code and resources for this work are publicly available at:
\url{https://github.com/Inamullah-Colab/ReitSEM}.
\end{abstract}

\section{Introduction}

Retinal imaging provides a unique, non-invasive window into systemic cardiovascular health, motivating its widespread use in modern AI-driven diagnostic systems~\cite{yuan2024retinal}. Recent deep learning approaches have demonstrated strong predictive performance for cardiovascular risk estimation directly from retinal images~\cite{poplin2018retina}. However, these models are predominantly associative and often operate as black-box predictors, offering limited insight into the underlying biological mechanisms or how upstream physiological variation may be reflected in retinal microvascular structure and related to downstream vascular outcomes~\cite{castro2020causality}. In particular, they provide limited explanation of how upstream physiological variation may be reflected in retinal microvascular structure and related to downstream vascular outcomes.

In this work, the retinal block is not interpreted as a direct biological cause of downstream cardiovascular or vascular outcomes. Instead, retinal traits are treated as an accessible phenotype layer that may statistically reflect upstream systemic vascular, metabolic, and inflammatory processes. The proposed mediator structure is therefore used as a hypothesis-testing strategy: it evaluates whether retinal microvascular features carry measurable mediator-like signal between systemic exposures and vascular outcomes under explicit modelling assumptions. This framing is consistent with the view of retinal microvascular signs as a non-invasive window onto systemic vascular dysfunction, rather than as isolated causal drivers of cardiovascular disease~\cite{wong2005systemic,mcgeechan2009retinal,wang2025ai}. Beyond prediction, a central scientific question is whether retinal traits behave as passive correlates, downstream biomarkers, or phenotype-level indicators that may carry mediator-like statistical signal between systemic exposures and vascular outcomes.

\begin{figure}[t]
\centering
\includegraphics[width=\linewidth]{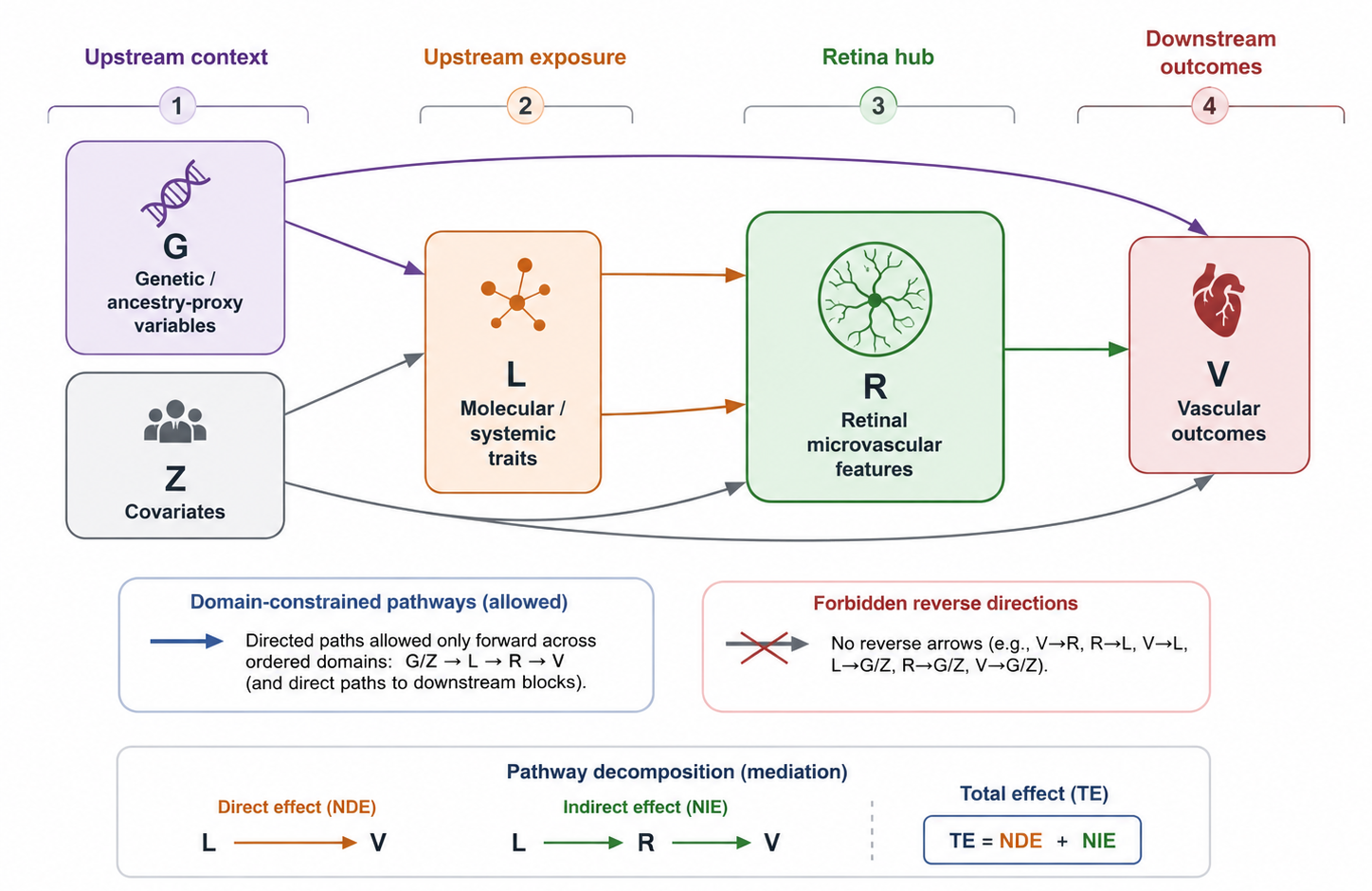}
\caption{
RetiSEM pathway prior. Genetic or ancestry-proxy variables ($G$), covariates ($Z$), and molecular traits ($L$) are organised upstream of retinal microvascular features ($R$) and vascular outcomes ($V$). Directed edges show the domain-constrained pathways allowed during structure learning and pathway decomposition.
}
\label{fig:dag}
\end{figure}

This question is difficult to address in real-world biomedical settings because multimodal data are often fragmented. Clinical variables, molecular measurements, and retinal imaging are typically collected in independent cohorts and are rarely available jointly at the participant level~\cite{acosta2022multimodal}. For example, NHANES provides extensive clinical and physiological information~\cite{nchs_nhanes_data}, while retinal imaging datasets such as APTOS 2019 are collected separately~\cite{aptos2019_data}. As a result, causal modelling must operate under limited-resource conditions, requiring harmonisation of partially observed modalities rather than assuming access to a complete participant-linked multimodal dataset.

Classical causal discovery methods, including PC~\cite{spirtes2000cps}, NOTEARS~\cite{zheng2018notears}, and LiNGAM~\cite{shimizu2006lingam}, provide principled frameworks for learning directed relationships from observational data. However, in high-dimensional and noisy biomedical settings, these approaches can produce unstable or biologically implausible structures and may generalise poorly across heterogeneous data regimes~\cite{pawlowski2020deep}. This motivates causal modelling frameworks that incorporate domain structure while retaining interpretable pathway-level analysis.

To address this gap, we propose \textbf{RetiSEM}, a domain-constrained structural equation modelling framework for fragmented multimodal biomedical data. RetiSEM enforces an ordered variable structure across genetics, molecular traits, retinal features, and vascular outcomes, while restricting implausible edges through a forbidden-edge masking mechanism. The resulting pathway prior is illustrated in Fig.~\ref{fig:dag}. 

\begin{figure}[t]
\centering
\includegraphics[width=\linewidth]{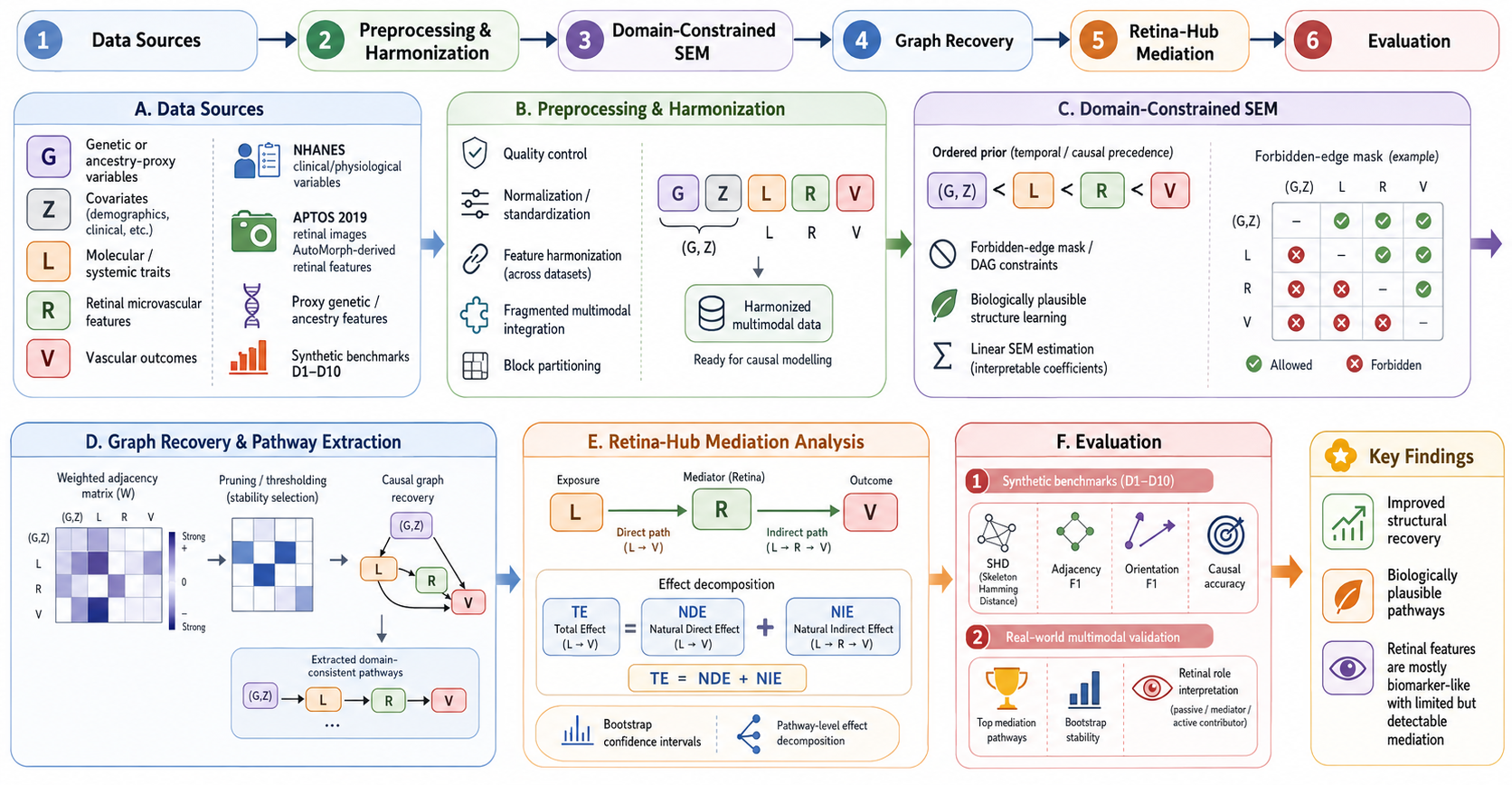}
\caption{
Overview of the RetiSEM workflow, from fragmented biomedical data harmonisation to constrained graph recovery, retina-hub mediation analysis, and evaluation.
}
\label{fig:retisem_workflow}
\end{figure}

RetiSEM is designed not only for structure learning but also for pathway-level interpretation. By incorporating causal mediation analysis, the framework decomposes estimated effects into total, direct, and indirect components, enabling systematic evaluation of whether retinal variables behave as passive indicators, reflective biomarkers, or mediator-like statistical components under explicit modelling assumptions. The overall workflow is shown in Fig.~\ref{fig:retisem_workflow}.

We validate the framework in both synthetic and real-world fragmented settings. The synthetic benchmark suite contains ten scenarios with varying dimensionality, nonlinearity, causal depth, and parallel pathway structure, allowing assessment of whether domain constraints improve graph recovery as the data-generating process becomes more complex. In the real-world analysis, NHANES clinical variables are combined with externally derived retinal representations and proxy population-structure variables, reflecting a low-resource setting in which biomedical modalities are not fully participant-linked.

The main contribution of this work is a domain-constrained causal modelling framework for disjoint multimodal biomedical data. RetiSEM combines biologically ordered variable blocks, forbidden-edge masking, structural equation modelling, and causal mediation analysis to recover interpretable pathway structures under limited-resource conditions. Collectively, the experiments evaluate whether domain constraints improve causal graph recovery and whether retinal phenotypes carry passive, reflective, or mediator-like pathway signal in vascular outcome modelling.

\section{Method}
\begin{table}[t]
\centering
\footnotesize
\caption{
Synthetic benchmark design (D1--D10) varying dimensionality ($p$), sample size ($n$), nonlinearity ($\rho_{\text{nonlin}}$), parallel paths ($\kappa$), and causal depth ($\ell$).
}
\label{tab:d1_d10_config_map}
\setlength{\tabcolsep}{3pt}
\begin{tabular}{l l c c c c c p{2.8cm}}
\toprule
ID & Dataset & $p$ & $n$ & $\rho$ & $\kappa$ & $\ell$ & Description \\
\midrule
D1  & LowDim-L  & 20  & 6000 & 0.0 & 1 & 3 & Linear baseline \\
D2  & LowDim-N  & 20  & 6000 & 0.5 & 1 & 3 & Nonlinear baseline \\
D3  & LowDim-P  & 20  & 6000 & 0.5 & 2 & 2 & Parallel paths \\
D4  & LowDim-D  & 20  & 6000 & 0.5 & 1 & 6 & Deep nonlinear chain \\
D5  & MidDim-C  & 50  & 6000 & 0.5 & 1 & 6 & Confounded regime \\
D6  & MidDim-P  & 100 & 6000 & 0.5 & 2 & 2 & Parallel nonlinear \\
D7  & MidDim-D  & 100 & 6000 & 0.5 & 1 & 6 & Deep causal chain \\
D8  & MidDim-S  & 100 & 6000 & 0.5 & 1 & 3 & Shallow nonlinear \\
D9  & HighDim-S & 200 & 6000 & 0.5 & 1 & 6 & High-dim deep \\
D10 & HighDim-D & 200 & 6000 & 0.5 & 1 & 3 & High-dim shallow \\
\bottomrule
\vspace{-0.8cm}
\end{tabular}
\end{table}


We propose a domain-constrained SEM framework for causal graph recovery and pathway-level interpretation in data-scarce multimodal biomedical settings. The central objective is to recover biologically consistent causal structure while characterising retinal variables as passive indicators, reflective biomarkers, or mediator-like statistical components within cardiovascular pathways. The overall causal hypothesis is illustrated in Fig.~\ref{fig:dag}, where upstream genetic and molecular factors propagate through retinal features before influencing vascular outcomes.

\subsection{Problem Setup and Domain Structure}

Let $X \in \mathbb{R}^{n \times d}$ denote the observed data matrix, where variables are partitioned into biologically meaningful blocks:
\[
(G, Z) \prec L \prec R \prec V,
\]
with $G$ denoting genetic or ancestry-proxy features, $Z$ covariates, $L$ molecular traits, $R$ retinal microvascular features, and $V$ vascular outcomes. This ordering defines a domain-informed statistical pathway prior in which retinal variables form an interpretable phenotype layer between upstream biological context and vascular outcome modelling. Within RetiSEM, this structure is used to evaluate whether retinal microvascular features carry mediator-like statistical signal, reflective biomarker information, or only weak associative signal under the specified assumptions. The imposed ordering constrains the DAG search space, reduces biologically implausible reverse-direction edges, and improves interpretability in limited-resource biomedical settings. Compared with unconstrained DAG learning, this design uses prior biomedical knowledge to guide structure recovery while retaining RetiSEM as a structured hypothesis-testing framework.

\subsection{Domain-Constrained Graph Recovery}

We aim to estimate a directed acyclic graph (DAG) represented by a weighted adjacency matrix $W \in \mathbb{R}^{d \times d}$. To prevent invalid relationships, we define a forbidden-edge mask $F \in \{0,1\}^{d \times d}$ such that:

\[
\hat{W} = W \odot (1 - F),
\]

where $F_{ij} = 1$ indicates disallowed edges (e.g., $V \rightarrow L$ or $R \rightarrow G$). This enforces strict adherence to the domain ordering $(G, Z) \prec L \prec R \prec V$. 

This constraint effectively reduces the hypothesis space, improving stability under limited data and preventing structurally invalid graph recovery. In particular, it ensures acyclicity by construction and avoids reverse-direction edges that frequently arise in unconstrained methods. The mask does not force every variable to participate in a single fixed path. Instead, it defines a permissible pathway 
space: edges consistent with the RetiSEM prior can be estimated when supported by the data, whereas reverse or biologically implausible directions are excluded. In this sense, the framework can follow the prior pathway organisation illustrated in Fig. \ref{fig:dag}, while still allowing the learned graph to select a sparse subset of supported directed relationships.

\subsection{Structural Equation Modelling and Estimation}

The pathway structure is modelled using SEM over the ordered domain blocks:
\[
L = f_L(G, Z) + \epsilon_L,
\]
\[
R = f_R(L, G, Z) + \epsilon_R,
\]
\[
V = f_V(R, L, G, Z) + \epsilon_V.
\]
Here, $L$ denotes molecular or systemic exposure variables, $R$ denotes retinal microvascular phenotype variables, $V$ denotes vascular outcomes, and $(G,Z)$ denote genetic or ancestry-proxy variables and covariates. This formulation allows upstream biological context to influence molecular traits, retinal phenotypes, and vascular outcomes while preserving the domain-informed ordering introduced above. In practice, we adopt a linear SEM approximation as an interpretable and statistically stable estimator for causal graph recovery and pathway-level effect decomposition. Although biomedical relationships may involve nonlinear and interaction-driven mechanisms, the linear approximation provides a tractable first-order model for estimating total, direct, and indirect components across high-dimensional and noisy biomedical datasets. This choice also enables consistent comparison across simulated and real-data settings, while keeping the estimated pathways transparent and reproducible.

Under this approximation, each structural component is estimated using regression-based models subject to the domain-constrained adjacency mask. The learned coefficients are then used to quantify pathway-specific effects through the mediation formulation described below.

\subsection{Retina-Hub Mediation Modeling}

The model uses causal mediation analysis to quantify how much pathway-level signal passes through the retinal phenotype block. For an upstream exposure $L$, retinal phenotype layer $R$, vascular outcome $V$, and adjustment variables $(G,Z)$, let $R(l)$ denote the retinal phenotype value under exposure level $l$, and let $V(l,r)$ denote the vascular outcome under exposure level $l$ and retinal state $r$. We write $V(l)=V(l,R(l))$ for the potential outcome when both the exposure and retinal phenotype take their natural values under $l$. Comparing exposure levels $l$ and $l^{\ast}$, the total effect (TE), natural indirect effect (NIE), and natural direct effect (NDE) are defined as:
\[
\mathrm{TE}(l,l^{\ast}) =
\mathbb{E}\{V(l)-V(l^{\ast})\},
\]
\[
\mathrm{NIE}(l,l^{\ast}) =
\mathbb{E}\{V(l,R(l))-V(l,R(l^{\ast}))\},
\]
\[
\mathrm{NDE}(l,l^{\ast}) =
\mathbb{E}\{V(l,R(l^{\ast}))-V(l^{\ast},R(l^{\ast}))\}.
\]

Under consistency, positivity, and conditional exchangeability assumptions, these quantities can be identified as pathway-level summaries of direct and indirect association components~\cite{imai2010general,vanderweele2015explanation}. Let $C=(G,Z)$ denote the adjustment set, and define
\[
\mu_V(l,r,c)=\mathbb{E}[V \mid L=l, R=r, C=c].
\]
The nested counterfactual mean is identified by the mediation functional:
\[
\begin{aligned}
\mathbb{E}\{V(l,R(l^{\ast}))\}
&=
\int \int
\mu_V(l,r,c)\,
p(r \mid L=l^{\ast}, C=c)\,
p(c)\,dr\,dc .
\end{aligned}
\]
This gives the standard decomposition:
\[
\mathrm{TE}(l,l^{\ast})
=
\mathrm{NDE}(l,l^{\ast})
+
\mathrm{NIE}(l,l^{\ast}).
\]

In the linear SEM implementation, these counterfactual quantities are operationalised through a product-of-coefficients estimator:
\[
R = \alpha L + \gamma^\top [G,Z] + \epsilon_R,
\]
\[
V = c'L + \beta R + \delta^\top [G,Z] + \epsilon_V.
\]
Here, $\alpha$ captures the exposure--retina association, $\beta$ captures the retina--outcome association conditional on $L$, $G$, and $Z$, and $c'$ captures the direct exposure--outcome component. For a single retinal mediator, the pathway components are estimated as
\[
\mathrm{NIE}=\alpha\beta, \qquad
\mathrm{NDE}=c', \qquad
\mathrm{TE}=c' + \alpha\beta.
\]
For a retinal block with multiple features $R_1,\ldots,R_m$, the indirect component is estimated by summing feature-level pathways:
\[
\mathrm{NIE}=\sum_{j=1}^{m}\alpha_j\beta_j, \qquad
\mathrm{TE}=c' + \sum_{j=1}^{m}\alpha_j\beta_j.
\]
This decomposition allows RetiSEM to distinguish weak associative signals from more stable mediator-like pathway contributions based on the magnitudes and robustness of the estimated direct and indirect components.

\begin{table*}[t]
\centering
\footnotesize
\caption{
Causal discovery performance on benchmark datasets.
(D1--D10) Best values per dataset are shown in bold. Higher is better for Adjacency F1,Orientation F1, and Causal Accuracy; lower is better for SHD.}
\label{tab:causal_discovery_benchmark}
\resizebox{\textwidth}{!}{%
\begin{tabular}{llcccccccccc}
\toprule
\textbf{Method} & \textbf{Metric} & \textbf{D1} & \textbf{D2} & \textbf{D3} & \textbf{D4} & \textbf{D5} & \textbf{D6} & \textbf{D7} & \textbf{D8} & \textbf{D9} & \textbf{D10} \\
\midrule

\multirow{4}{*}{PC~\cite{spirtes2000cps}}
& SHD & 0.3262 & 0.3006 & 0.3333 & 0.3034 & 0.3183 & 0.3329 & 0.3225 & 0.3246 & 0.3218 & 0.3231 \\
& Adjacency F1 & 0.3108 & 0.3878 & 0.4306 & 0.3813 & 0.2175 & 0.1025 & 0.1099 & 0.1089 & 0.0332 & 0.0342 \\
& Orientation F1 & 0.0876 & 0.1977 & 0.1875 & 0.1712 & 0.0558 & 0.0177 & 0.0193 & 0.0171 & 0.0058 & 0.0056 \\
& Causal Accuracy & 0.6738 & 0.6994 & 0.6667 & 0.6966 & 0.6817 & 0.6671 & 0.6775 & 0.6754 & 0.6782 & 0.6769 \\
\midrule

\multirow{4}{*}{NOTEARS~\cite{zheng2018notears}}
& SHD & 0.2379 & 0.2735 & 0.3319 & 0.2479 & 0.2860 & 0.2965 & 0.2965 & 0.2899 & 0.3365 & 0.3371 \\
& Adjacency F1 & 0.2918 & 0.1273 & \textbf{0.7509} & 0.1683 & 0.0379 & 0.1092 & 0.1092 & 0.0759 & \textbf{0.6685} & \textbf{0.6768} \\
& Orientation F1 & 0.2833 & 0.1273 & \textbf{0.5738} & 0.1386 & 0.0379 & 0.1087 & 0.1087 & 0.0759 & 0.4863 & 0.4888 \\
& Causal Accuracy & 0.7621 & 0.7265 & 0.6296 & 0.7521 & 0.7140 & 0.7035 & 0.7035 & 0.7101 & 0.5575 & 0.5595 \\
\midrule

\multirow{4}{*}{DAGMA \cite{bello2022dagma}}
& SHD & 0.2920 & 0.3262 & 0.3319 & 0.3490 & 0.3819 & 0.3738 & 0.3807 & 0.3750 & 0.3365 & 0.3371 \\
& Adjacency F1 & 0.6931 & \textbf{0.7048} & \textbf{0.7509} & 0.6844 & 0.6838 & \textbf{0.7130} & \textbf{0.7105} & 0.6885 & \textbf{0.6685} & \textbf{0.6768} \\
& Orientation F1 & 0.5599 & 0.5681 & \textbf{0.5738} & 0.4783 & 0.4685 & 0.4837 & 0.4772 & 0.4640 & 0.4863 & 0.4888 \\
& Causal Accuracy & 0.6282 & 0.6339 & 0.6296 & 0.5712 & 0.5486 & 0.5545 & 0.5488 & 0.5416 & 0.5575 & 0.5595 \\
\midrule

\multirow{4}{*}{LiNGAM \cite{shimizu2006lingam}}
& SHD & 0.4430 & 0.3476 & 0.3917 & 0.3419 & 0.4239 & 0.4575 & 0.4393 & 0.4501 & 0.4019 & 0.4083 \\
& Adjacency F1 & 0.5681 & 0.4311 & 0.4654 & 0.5094 & 0.5241 & 0.5065 & 0.5328 & 0.5183 & 0.3360 & 0.3411 \\
& Orientation F1 & 0.0986 & 0.2695 & 0.2382 & 0.2453 & 0.1069 & 0.0449 & 0.0532 & 0.0300 & 0.0079 & 0.0100 \\
& Causal Accuracy & 0.5570 & 0.6524 & 0.6083 & 0.6581 & 0.5761 & 0.5425 & 0.5607 & 0.5499 & 0.5981 & 0.5917 \\
\midrule

\multirow{4}{*}{DECI \cite{geffner2022deci}}
& SHD & 0.3405 & 0.3262 & 0.3319 & 0.3490 & 0.3816 & 0.3812 & 0.3856 & 0.3738 & 0.3365 & 0.3371 \\
& Adjacency F1 & 0.7234 & \textbf{0.7048} & \textbf{0.7509} & 0.6844 & 0.6797 & 0.7108 & 0.6845 & \textbf{0.6908} & \textbf{0.6685} & \textbf{0.6768} \\
& Orientation F1 & 0.5105 & 0.5681 & \textbf{0.5738} & 0.4783 & 0.4677 & 0.4757 & 0.4465 & 0.4652 & 0.4863 & 0.4888 \\
& Causal Accuracy & 0.5356 & 0.6339 & 0.6296 & 0.5712 & 0.5479 & 0.5476 & 0.5272 & 0.5427 & 0.5575 & 0.5595 \\
\midrule

\multirow{4}{*}{RetiSEM}
& SHD & \textbf{0.1211} & \textbf{0.1524} & \textbf{0.1838} & \textbf{0.1140} & \textbf{0.1369} & \textbf{0.1753} & \textbf{0.1747} & \textbf{0.1658} & \textbf{0.1851} & \textbf{0.1829} \\
& Adjacency F1 & \textbf{0.7267} & 0.6298 & 0.5536 & \textbf{0.7059} & \textbf{0.7005} & 0.6037 & 0.5967 & 0.6252 & 0.5772 & 0.5863 \\
& Orientation F1 & \textbf{0.7267} & \textbf{0.6298} & 0.5536 & \textbf{0.7059} & \textbf{0.7005} & \textbf{0.6037} & \textbf{0.5967} & \textbf{0.6252} & \textbf{0.5772} & \textbf{0.5863} \\
& Causal Accuracy & \textbf{0.8789} & \textbf{0.8476} & \textbf{0.8162} & \textbf{0.8860} & \textbf{0.8631} & \textbf{0.8247} & \textbf{0.8253} & \textbf{0.8342} & \textbf{0.8149} & \textbf{0.8171} \\
\bottomrule
\end{tabular}%
}
\end{table*}

\subsection{Fragmented Multimodal Evaluation}

To reflect real-world biomedical constraints, RetiSEM is evaluated under both controlled synthetic regimes and a fragmented applied setting. In the applied analysis, clinical and physiological variables are obtained from NHANES, retinal representations are derived from external fundus-image resources, and proxy genetic or ancestry-related variables are used for population-structure adjustment where individual genotype data are unavailable. Retinal representations are extracted using the AutoMorph~\cite{zhou2022automorph} retinal image-analysis pipeline, producing quantitative microvascular descriptors such as vessel calibre, tortuosity, vessel density, and branching-related morphology, which define the retinal phenotype block $R$.

\begin{figure}[b!]
\centering
\includegraphics[width=\linewidth]{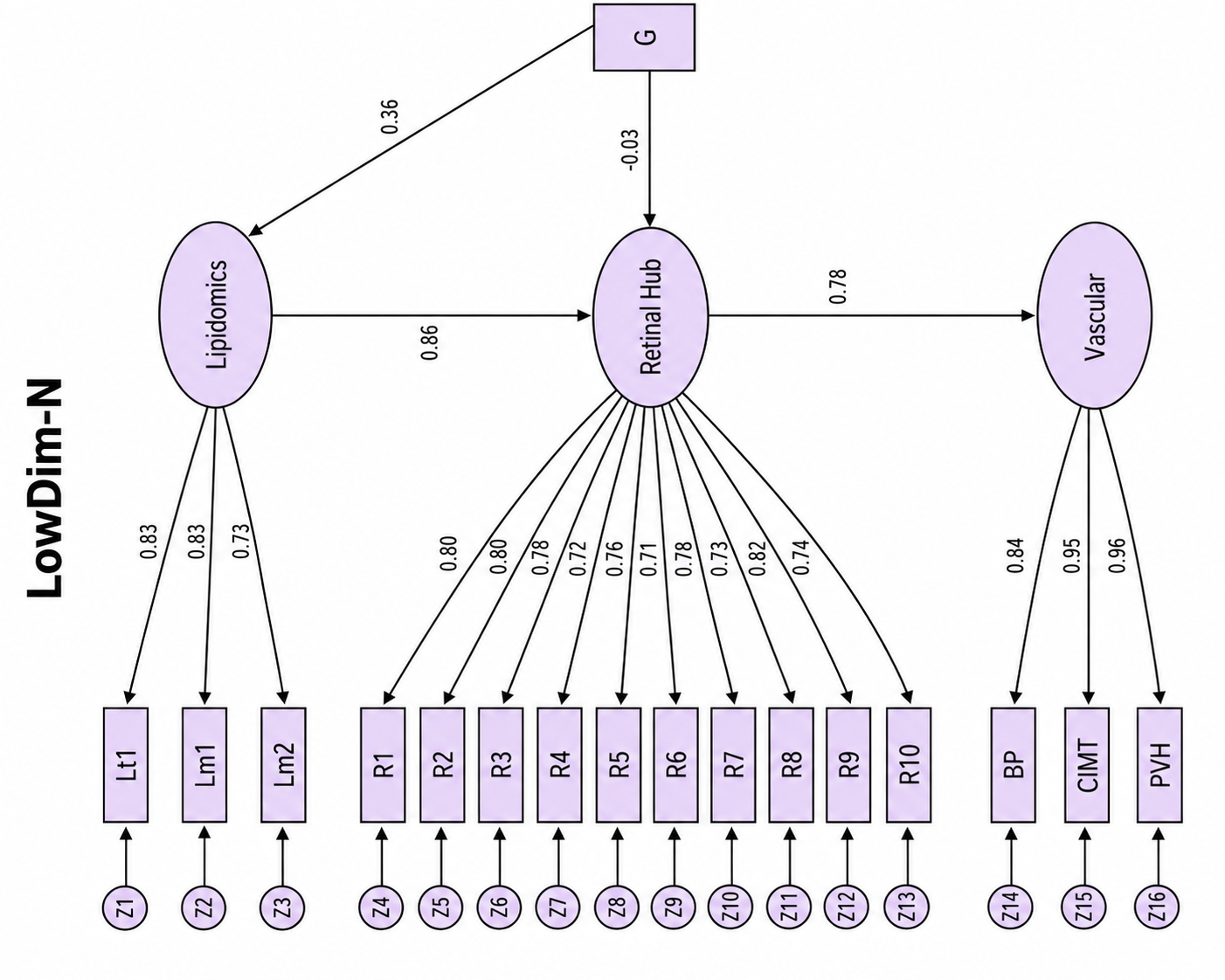}
\caption{
Illustrative mediation pathway structure for the LowDim-N synthetic setting. The diagram shows how RetiSEM organises lipidomic, retinal, and vascular indicators into a constrained pathway model, with path coefficients shown for the representative low-dimensional regime. Full TE, NDE, and NIE decompositions across all synthetic benchmark scenarios are reported in Table~\ref{tab:manifest_paths_d1_d10}.}
\label{fig:mediation}
\end{figure}

The ten synthetic regimes are generated using the block-structured mechanism summarised in Table~\ref{tab:d1_d10_config_map}. Each regime follows the ordered pathway prior $(G,Z) \prec L \prec R \prec V$, while varying dimensionality, nonlinearity, causal depth, and parallel pathway structure to test the framework under progressively more complex generative conditions. Fig.~\ref{fig:mediation} illustrates one representative LowDim-N pathway structure, showing how lipidomic, retinal, and vascular feature blocks are organised within the constrained mediation design.

This evaluation design tests whether the imposed pathway prior remains informative when moving from known synthetic structures to realistic data fragmentation. Although estimation uses a linear SEM approximation for interpretability and identifiability, the experiments assess whether the constrained pathway design remains useful across nonlinear synthetic regimes and partially observed multimodal data.

\section{Experiments}

We evaluated the proposed framework on both synthetic and real-world multimodal datasets to assess its ability to recover causal structure and characterize retinal-mediated pathways. Our experimental design focuses on three key aspects: (i) structural recovery accuracy, (ii) robustness under diverse generative conditions, and (iii) interpretation of retinal variables within cardiovascular pathways.
\subsection{Synthetic Benchmark Evaluation}

To systematically evaluate generalization, we construct a suite of ten synthetic benchmark scenarios (D1–D10) that vary along multiple axes, including dimensionality (low, medium, high), causal depth (shallow vs.\ deep), and functional form (linear vs.\ nonlinear). Each scenario follows the biologically motivated ordering $(G,Z) \prec L \prec R \prec V$, ensuring consistency with the domain assumptions of RetiSEM.
\begin{table}[t]
\centering
\scriptsize
\caption{
Representative mediation chains and linear-SEM effect decomposition (TE, NDE, NIE) for synthetic scenarios D1--D10.
}
\label{tab:manifest_paths_d1_d10}
\setlength{\tabcolsep}{2pt}
\begin{tabular}{l >{\raggedright\arraybackslash}p{3.8cm} c c c}
\toprule
ID & Chain & TE & NDE & NIE \\
\midrule
D1  & Lt1$\rightarrow$Lm1$\rightarrow$R1$\rightarrow$Vbp & 0.0469 & -0.0062 & 0.0245 \\
D2  & Lt1$\rightarrow$Lm1$\rightarrow$R1$\rightarrow$Vbp & 0.0643 & -0.0073 & 0.0585 \\
D3  & Lt1$\rightarrow$Zfix\_PC1$\rightarrow$Vbp & 0.2168 & 0.1745 & -0.0050 \\
D4  & Lt1$\rightarrow$R4$\rightarrow$Vbp & 0.0121 & -0.0157 & -0.0150 \\
D5  & Lt3$\rightarrow$Lm5$\rightarrow$R6$\rightarrow$Vpvh & -0.0007 & 0.0269 & -0.0611 \\
D6  & Lt8$\rightarrow$Lm7$\rightarrow$R44$\rightarrow$Vbp\_pvh & 0.3513 & 0.3740 & 0.0188 \\
D7  & Lt14$\rightarrow$Lm14$\rightarrow$R10$\rightarrow$Vpvh & 0.4707 & 0.2012 & 0.0095 \\
D8  & Lt8$\rightarrow$Lm10$\rightarrow$R16$\rightarrow$Vbp\_cimt\_pvh & -0.0769 & -0.0073 & -0.1027 \\
D9  & Lt20$\rightarrow$Lm16$\rightarrow$R58$\rightarrow$Vbp\_pvh & 0.3141 & 0.6304 & 1.1465 \\
D10 & Lt40$\rightarrow$Lm21$\rightarrow$R62$\rightarrow$Vbp\_cimt\_pvh & 2.6264 & 0.0173 & 3.2950 \\
\bottomrule
\end{tabular}
\end{table}

\paragraph{Synthetic Scenario Design.}
The synthetic benchmarks are designed to systematically stress-test the proposed framework across increasing levels of complexity. As summarized in Table~\ref{tab:d1_d10_config_map}, the scenarios vary along three primary axes: (i) functional complexity through nonlinear transformations ($\rho_{\text{nonlin}}$), (ii) structural complexity via parallel paths ($\kappa$) and causal depth ($\ell$), and (iii) scalability through increasing dimensionality ($p$). This structured design ensures that the evaluation does not rely on idealized assumptions. In particular, the inclusion of both linear (D1) and nonlinear scenarios (D2–D10) allows us to assess whether RetiSEM remains effective when the underlying data-generating mechanisms deviate from linearity.
These benchmarks are designed to test whether the proposed framework can reliably recover graph structure and pathway-level relationships under increasingly challenging conditions.
Table~\ref{tab:causal_discovery_benchmark} reports the full quantitative comparison across all ten synthetic benchmark scenarios. Across D1-D10, RetiSEM 
achieves the lowest SHD and the highest causal accuracy in every setting, indicating more accurate recovery of the underlying graph structure under both simple and challenging regimes. Some baseline methods achieve strong adjacency F1 in selected scenarios, but their performance is less stable across metrics and regimes. Collectively, RetiSEM provides the 
most consistent balance between low structural error, reliable orientation recovery, and high causal accuracy across the benchmark suite.

Beyond graph recovery, we examine whether the recovered structures preserve meaningful pathway-level effects. Figure~\ref{fig:mediation} provides an illustrative example of the mediation pathway structure in the LowDim-N synthetic setting. Because the full mediation outputs across all ten scenarios are difficult to visualise in a single compact figure, this representative low-dimensional case shows how RetiSEM organises lipidomic, retinal, and vascular indicator blocks into a constrained 
pathway model. The displayed coefficients indicate loading and pathway strengths within this example, while the corresponding TE, NDE, and NIE decompositions across all benchmark scenarios are reported in Table~\ref{tab:manifest_paths_d1_d10}.

\begin{figure*}[t]
\centering
\includegraphics[width=0.97\textwidth,height=0.60\textheight,keepaspectratio]{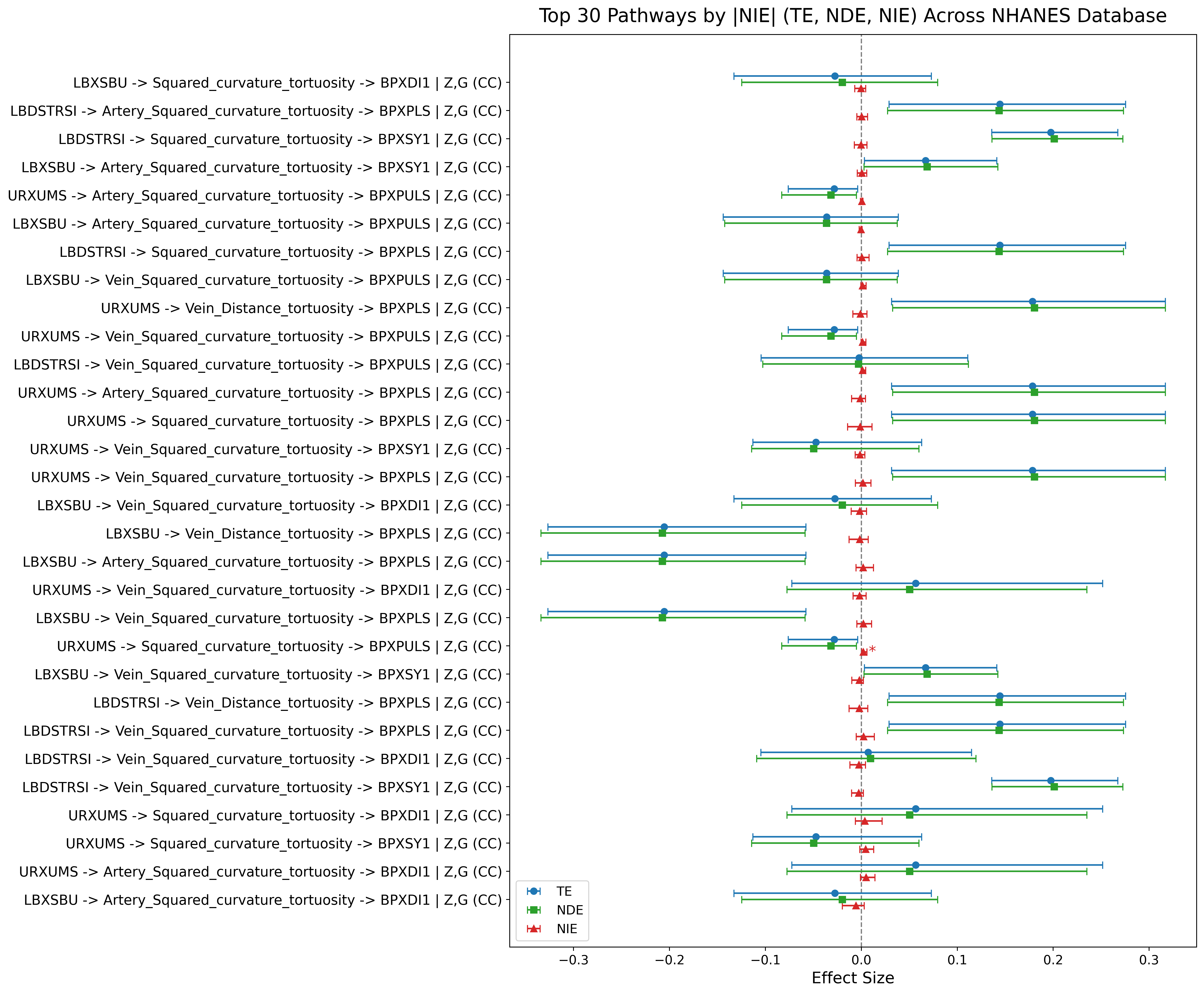}
\vspace{-0.5em}
\caption{Top NHANES mediation pathways ranked by $|\mathrm{NIE}|$. Blue circles denote TE, green squares denote NDE, and red triangles denote NIE.}
\label{fig:realdata}
\vspace{-0.5em}
\end{figure*}

\subsection{Real-World Multimodal Analysis}

To evaluate RetiSEM under realistic fragmentation, we combine NHANES clinical and physiological variables with externally derived retinal representations and proxy population-structure variables. This setting does not assume complete participant-level multimodal alignment; instead, it tests whether a domain-constrained causal framework can operate when biomedical modalities are partially observed and harmonised through feature definitions and covariate structure.

Across 96 exposure--retinal mediator--vascular outcome combinations, TE and NDE dominate most relationships, whereas NIE values are generally smaller but non-zero. Figure~\ref{fig:realdata} ranks the strongest NHANES-based pathways by $|\mathrm{NIE}|$ to identify where the retinal block contributes the largest mediator-like statistical signal. Each pathway follows the adjusted complete-case specification $L \rightarrow R \rightarrow V \mid Z,G$ (CC), where $L$ denotes an upstream exposure, $R$ denotes a retinal mediator, and $V$ denotes a vascular outcome.

For example, \texttt{LBDSTRSI} (refrigerated serum triglycerides, mmol/L) $\rightarrow$ 
\texttt{Vein\_Distance\_tortuosity} $\rightarrow$ \texttt{BPXPLS} can be interpreted as a lipid-related systemic exposure associated with a vascular examination outcome through a retinal tortuosity feature, after adjustment for $Z$ and $G$. This example illustrates how the model reports both the direct pathway component and the retinal indirect component within the same adjusted specification. The overall pattern in Fig.~\ref{fig:realdata} shows that even among the highest-ranked indirect pathways, NIE remains smaller than TE and NDE in most cases. This supports the interpretation that retinal traits mainly behave as downstream vascular-reflective biomarkers in this cohort, while retaining limited but detectable mediator-like statistical contributions in selected pathways.

\subsection{Analysis of Retinal Roles}

The mediation results support the interpretation of retinal variables along a continuum rather than as fixed biological categories. Features with negligible indirect effects are best understood as weak associative indicators. Features with stable non-zero NIE values provide a mediator-like statistical signal under the specified pathway assumptions. Features with stronger direct associations with vascular outcomes may serve as reflective downstream biomarkers of systemic vascular state. Across both synthetic and real-data analyses, most retinal variables fall closer to the biomarker or weak-mediator end of this continuum, reinforcing the view that retinal morphology reflects systemic vascular variation more strongly than it acts as a dominant causal driver of downstream vascular outcomes.
Overall, the experiments show that RetiSEM improves structural recovery in synthetic settings with increasing dimensionality, nonlinearity, and causal depth, while preserving interpretable TE/NDE/NIE decomposition. The real-data analysis further demonstrates how the framework can be applied under fragmented multimodal resources, where retinal traits provide mainly biomarker-like information with smaller but measurable indirect effects. These findings support the use of domain-constrained causal priors 
as a practical strategy for interpretable pathway modelling in limited-resource biomedical AI.

\section{Discussion}

The results show that domain-constrained structure learning can improve causal graph recovery in fragmented multimodal biomedical settings. Table \ref{tab:d1_d10_config_map} defines a benchmark suite that varies dimensionality, nonlinearity, causal depth, and parallel 
pathway structure, providing a stress test beyond a single idealised simulation. The quantitative results in Table \ref{tab:causal_discovery_benchmark} show 
that RetiSEM achieves the lowest SHD and highest causal accuracy across D1-D10. This suggests that biologically informed masks can reduce unstable or implausible edge orientations when data are limited, noisy, or generated under heterogeneous regimes. The mediation results further show that RetiSEM provides more than edge-level graph recovery. Fig. \ref{fig:mediation} illustrates how the LowDim-N setting is represented as a constrained pathway model, while Table \ref{tab:manifest_paths_d1_d10} reports TE, NDE, and NIE decompositions across all synthetic scenarios. Together, these results support the methodological claim that domain priors 
and SEM-based decomposition can be combined to evaluate whether a retinal phenotype layer carries pathway-level statistical signal, even when the underlying generative process is not perfectly linear. In the real-world analysis, Figure \ref{fig:realdata} indicates that indirect effects through retinal variables are generally smaller than total 
and direct effects. This pattern prevents over-interpretation of the retina as a dominant causal mediator. Instead, the findings suggest that retinal microvascular features behave primarily as downstream biomarker-like indicators of systemic vascular variation, with selected pathways showing detectable mediator-like statistical signal. This interpretation is consistent with oculomics work that views retinal morphology as a non-invasive window into systemic circulation and cardiometabolic status \cite{wang2025ai}.

Several limitations should guide interpretation. The real-data setting remains an observational, low-data regime, and the retinal representations are externally derived rather than fully participant-linked to all NHANES variables. The linear SEM 
estimator provides interpretability and stability but cannot capture all nonlinear or interaction-driven biomedical 
mechanisms. Estimated indirect effects should therefore be viewed as pathway-level statistical summaries under structural 
assumptions, not as definitive biological proof. Future work should evaluate RetiSEM in participant-linked multimodal 
cohorts, extend the estimator to nonlinear structural models, and test the stability of inferred pathways across independent 
biomedical datasets.

\section{Conclusion}

In this work, we introduced RetiSEM, a domain-constrained structural equation modelling framework for 
causal graph recovery and mediation analysis under limited biomedical resources. By combining ordered biomedical variable blocks, forbidden-edge masking, and TE/NDE/NIE effect decomposition, RetiSEM supports interpretable pathway modelling when complete multimodal data are unavailable. Synthetic experiments across ten benchmark scenarios demonstrate improved structural recovery and interpretable mediation estimates under varying dimensionality, nonlinearity, and causal depth. The real-world analysis further suggests that retinal features behave mainly as downstream biomarker-like indicators, with a smaller but detectable mediator-like pathway signal. Overall, RetiSEM provides a practical and reproducible framework for testing structured causal hypotheses with limited resources in multimodal health data.

\bibliographystyle{named}
\bibliography{IJCAI26_GLOW}

\end{document}